\pdfoutput=1
\documentclass[sigplan,10pt,nonacm]{acmart}

\setcopyright{none}
\acmISBN{}
\acmDOI{}

\usepackage{hyperref}       %
\usepackage{url}            %
\usepackage{xurl}           %
\usepackage{booktabs}       %
\usepackage{nicefrac}       %
\usepackage{microtype}      %
\usepackage{xcolor}         %

\usepackage{listings}
\definecolor{keywordcolor}{rgb}{0.7, 0.1, 0.1}   %
\definecolor{tacticcolor}{rgb}{0.0, 0.1, 0.6}    %
\definecolor{commentcolor}{rgb}{0.4, 0.4, 0.4}   %
\definecolor{symbolcolor}{rgb}{0.0, 0.1, 0.6}    %
\definecolor{sortcolor}{rgb}{0.1, 0.5, 0.1}      %
\definecolor{attributecolor}{rgb}{0.7, 0.1, 0.1} %

\colorlet{punct}{red!60!black}
\definecolor{background}{HTML}{FFFFFF}
\definecolor{delim}{RGB}{20,105,176}
\colorlet{numb}{magenta!60!black}

\lstdefinelanguage{json}{
    basicstyle=\small\ttfamily,
    numbers=left,
    numberstyle=\scriptsize,
    stepnumber=1,
    numbersep=8pt,
    showstringspaces=false,
    breaklines=true,
    frame=lines,
    backgroundcolor=\color{background},
    literate=
     *{0}{{{\color{numb}0}}}{1}
      {1}{{{\color{numb}1}}}{1}
      {2}{{{\color{numb}2}}}{1}
      {3}{{{\color{numb}3}}}{1}
      {4}{{{\color{numb}4}}}{1}
      {5}{{{\color{numb}5}}}{1}
      {6}{{{\color{numb}6}}}{1}
      {7}{{{\color{numb}7}}}{1}
      {8}{{{\color{numb}8}}}{1}
      {9}{{{\color{numb}9}}}{1}
      {:}{{{\color{punct}{:}}}}{1}
      {,}{{{\color{punct}{,}}}}{1}
      {\{}{{{\color{delim}{\{}}}}{1}
      {\}}{{{\color{delim}{\}}}}}{1}
      {[}{{{\color{delim}{[}}}}{1}
      {]}{{{\color{delim}{]}}}}{1},
}

\usepackage{caption}
\usepackage{subcaption}
\usepackage{graphicx}
\usepackage{float}
\usepackage{tikz}
\usetikzlibrary{arrows.meta, positioning, calc, shapes, backgrounds}
\usepackage[export]{adjustbox}
\usepackage{enumitem}
\usepackage{tcolorbox}
\usepackage{multirow}

\usepackage{amsmath}
\usepackage{mathtools}
\usepackage{amsthm}

\theoremstyle{plain}

\theoremstyle{definition}

\theoremstyle{remark}

\usepackage{algorithm}
\usepackage{algorithmic}
\renewcommand{\algorithmicdo}{}
\renewcommand{\algorithmicthen}{}

\newenvironment{citemize}{\begin{list}{\scalebox{0.7}{$\bullet$}}{\topsep=.2\smallskipamount\itemsep=0pt\parsep=1pt\labelwidth=.5em}}{\end{list}}

\title[NanoProof: Open and Efficient Automated Theorem Proving in Lean 4]{NanoProof: Open and Efficient Automated\texorpdfstring{\\}{ }Theorem Proving in Lean 4}

\author[Matěj Kripner and Milan Straka]{Matěj Kripner\footnotemark[1]\hspace{2em}Milan Straka}
\makeatletter
\g@addto@macro\@authornotes{\stepcounter{footnote}\footnotetext{Corresponding author: \href{mailto:kripner@ufal.mff.cuni.cz}{kripner@ufal.mff.cuni.cz}.}}
\makeatother
\affiliation{%
  \institution{Institute of Formal and Applied Linguistics, Faculty of Mathematics and Physics, Charles University}
  \city{Prague}
  \country{Czech Republic}
}

\begin{abstract}
We introduce \textit{NanoProof}, to our knowledge the first factorized execution-guided theorem prover in Lean~4 whose training data, extraction tooling, training pipeline, and weights are all released, making it end-to-end reproducible using open-source resources.
To this end, we build and release a dataset of structured proof trees, as well as a tool for programmatic interaction and data extraction within the Lean~4 formal verifier.
To support sustainable research, we focus on compute efficiency to facilitate accessible training and evaluation.
NanoProof achieves 50.8\% pass@16 on MiniF2F-Test, exceeding the two closest systems of its class, HyperTree Proof Search and ABEL, at roughly $90\times$ and $7\times$ less compute, and using more than four orders of magnitude less compute than AlphaProof.
Stronger open-weight provers exist, but they are fine-tuned from large pretrained language models and release neither training data nor pipeline; NanoProof shows that the factorized execution-guided class of provers can be rebuilt from scratch with modest resources.
\end{abstract}

\begin{document}

\maketitle

\section{Introduction}

In automated theorem proving (ATP), the task at hand consists of generating a valid proof to a~mathematical statement.
We focus our attention on formal ATP, where both the statement and the proof are expressed in a formal language such as Lean~4~\citep{moura2021lean}.
In this setting, the correctness of a generated proof can be verified automatically, providing clear feedback to an ATP system, which can be used for training using reinforcement learning (RL).

Formal methods have found applications in the formalization of mathematics~\citep{doorn2023sphere_eversion}, physics~\citep{tooby_smith2024formalization_physics}, and chemistry~\citep{bobbin2024chemistry}, as well as software verification~\citep{avigad2025cairo_zero}, cryptography~\citep{doussot2024cryptography}, and general reasoning~\citep{jiang2024leanreasoner}.
With the current and anticipated rise of AI-generated code, proofs, and other artifacts, we expect human verification to become the bottleneck to innovation, making formal methods essential for their ability to automatically filter out incorrect solutions.
For example, mathematicians already benefit from expediting the review process in frontier research by submitting a formalization of their results~\citep{hales2024formal}.
Similarly, formal verification enables massively collaborative mathematical projects without the need for extensive human moderation~\citep{bolan2025equational}.

On an implementation level, ATP systems can be divided along several axes.
First, in an \textit{execution-guided} proof search, a prover interleaves proof search with verification, leveraging feedback from the formal verifier after each proof step and generating steps conditioned on the true internal proof state.
For example, in Lean~4, an execution-guided prover observes the list of currently open goals and is informed on any proof steps that resulted in an error.
In contrast, \textit{execution-agnostic} provers only use the formal verifier to verify the final proof, generating proof steps with no feedback about the validity of preceding steps or the current internal proof state.
Thus, in execution-agnostic search, it falls upon the proof step generator to deduce the evolving proof state, as this is a necessary precondition to determine the validity of subsequent proof steps.
This coupling of a policy model and a world model increases the complexity of proof generation.

Second, in a \textit{factorized} proof search, the prover is allowed to split proof states into separate branches that can be explored independently.
An illustrative example is mathematical induction, where factorized search considers the base case and the inductive case independently, enabling parallelization and dynamic allocation of search budget between the two branches.
The utility of node re-use and parallelization enabled by factorization has been verified by \citet{lample2022hypertree} and \citet{alphaproof} and we do not reiterate it in this work.
In contrast, in a~\textit{linear} proof search, the proof state consists of all the context necessary to complete the proof.

Finally, in an \textit{any-time} proof search, the prover performance grows when time or compute budget is increased, while a \textit{one-shot} prover can only be scaled by running multiple instances in parallel, without the possibility of utilizing results across searches.

Proof search can be treated as a generic text-to-text language modeling problem of generating a~proof given a statement, resulting in a linear one-shot execution-agnostic method.
Such methods have directly benefited from the recent rapid rise in capabilities of large language models (LLMs).
In contrast, factorized any-time execution-guided provers require careful pre-processing of training data and tight integration with the verification framework.
The existing systems of this class, HyperTree Proof Search~\citep{lample2022hypertree}, ABEL~\citep{gloeckle2024abel}, and AlphaProof~\citep{alphaproof}, released neither their data nor runnable code.
Open execution-guided provers such as InternLM2.5-StepProver~\citep{wu2024internlm2}, BFS-Prover~\citep{xin2025bfsprover}, and BFS-Prover-V2~\citep{xin2025bfsproverv2} perform linear best-first search over monolithic proof states and release at most model weights.
The factorized class has therefore been out of reach for the open-source community, both in terms of data availability and compute requirements.
We identify this as a major roadblock to advancing Machine Learning-based ATP in the age of large models, particularly since recent approaches like AlphaProof~\citep{alphaproof} and HyperTree Proof Search~\citep{lample2022hypertree} demonstrated the viability of this paradigm.

Since the inception of this work, text-to-text and agentic provers built on large language models have become extremely strong, with Seed-Prover~\citep{chen2025seedprover} and Goedel-Prover-V2~\citep{lin2025goedelv2} saturating MiniF2F.
We do not take this as evidence that search-based approaches are obsolete.
Casting proving as plain text generation means that a frontier LLM can be applied without adaptation, whereas factorized search-based provers require structured training data, verifier integration, and search infrastructure before they can run at all, and until now this machinery existed only inside closed systems.
Whether classical search with small models remains competitive with large generic LLMs at equal compute is an open question, and it can only be studied if this class of system exists in an open and reproducible form.
Moreover, the strongest approach may turn out to be a combination of the two, LLM reasoning with verifier-grounded search, a direction that the heavy use of Lean feedback in Seed-Prover already explores.
Factorized search also offers concrete advantages independent of model size: node re-use across branches, independent allocation of search budget to subgoals, and distance-to-proof-end labels that directly supervise the critic.
This is why we consider open, reproducible tooling for the search-based side worth building.

In this work, we address this reproducibility barrier by introducing \textit{NanoProof}, an end-to-end ATP system that leverages existing open resources and provides the novel components necessary to complete the Lean~4 pipeline.
Specifically, our contributions are as follows.
\begin{itemize}
\item \textbf{Data, Tooling, and Environment}~~We release LeanTree, a dataset of $92{,}927$ independently verified structured proof trees extracted from Mathlib, the extraction tool that produces such trees from any Lean~4 project, and a Lean interface with incremental kernel verification and a distributed environment pool for reinforcement learning.
\item \textbf{ATP System}~~We introduce a factorized any-time execution-guided prover called NanoProof and release all the tooling necessary to reproduce it and build on top of it.\footnote{The Python package and its source code is available at \url{https://github.com/kripner/nanoproof}.}
\item \textbf{Performance Evaluation}~~We demonstrate the efficacy of NanoProof through an end-to-end training run, achieving 50.8\% pass@16 on MiniF2F-Test~\citep{zheng2021minif2f} using more than four orders of magnitude less compute than AlphaProof.
\item \textbf{Search-Level Techniques}~~We describe practical train-time and inference-time techniques, namely proof simplification and augmentation, cycle detection, and diverse tactic sampling, and quantify their effect where our compute budget allowed.
\end{itemize}

In Section~\ref{sec:nanoproof}, we present the training pipeline and the search algorithm of NanoProof.
In Section~\ref{sec:leantree}, we describe the algorithm for extracting proof trees from human-authored Lean projects and the resulting LeanTree dataset, a necessary component for pre-training a factorized execution-guided prover.
Finally, in Section~\ref{sec:experiments}, we present a~training run of NanoProof using openly available datasets and LeanTree.
The necessary background in Lean and proof search is available in Appendix~\ref{sec:prelims}.

\begin{figure*}[t]
\centering
\includegraphics[width=\linewidth]{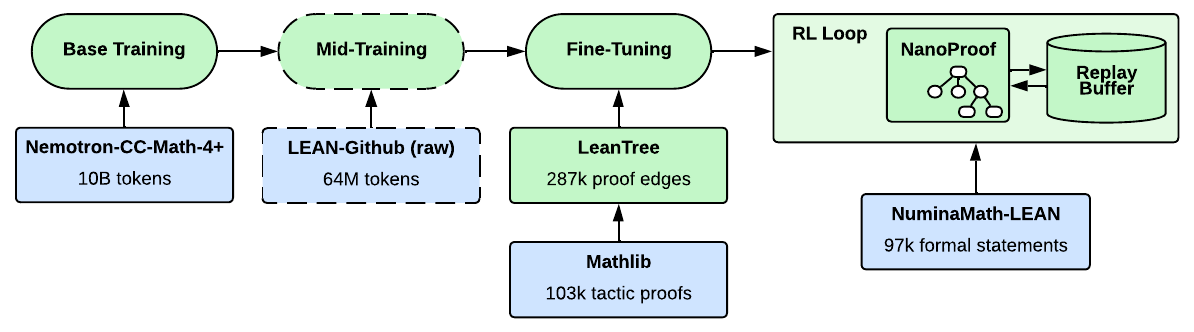}
\caption{
\textbf{NanoProof training pipeline.} After supervised pretraining, the prover enters into a \textit{self-improvement loop} where successfully discovered proofs are used for further training. Mid-training is not included in the final run, as described in Section~\ref{sec:experiments}.
}
\label{fig:overview}
\end{figure*}

\section{Related Work}\label{sec:related_work}

We give a brief overview of Lean, different ATP approaches, and existing formal ATP datasets and benchmarks.

\textbf{Lean 4}~\citep{moura2021lean} is a leading formal language for theorem proving with applications both in the formalization of mathematics~\citep{doorn2023sphere_eversion} and in numerous other domains~\citep{tooby_smith2024formalization_physics,bobbin2024chemistry,doussot2024cryptography,jiang2024leanreasoner}.
One of the appealing properties of Lean is its extensibility, stemming from the fact that its theorem proving component is written in Lean itself, and the user can modify both its syntax and its semantics, which we heavily utilize in LeanTree.

\textbf{Execution-Agnostic ATP.}
ATP can be formulated as a text-to-text task so that any existing LLM is directly applicable.
This approach to theorem proving was pioneered by Baldur~\citep{first2023baldur}.
More recently, LLMs using chain-of-thought refined using RL were utilized by DeepSeek-Prover-V2~\citep{ren2025deepseek}, Kimina-Prover~\citep{wang2025kimina}, InternLM-Math~\citep{ying2024internlm}, Goedel-Prover-V2~\citep{lin2025goedelv2}, and Seed-Prover~\citep{chen2025seedprover}, among others, securing top positions on relevant benchmarks; Goedel-Prover-V2 reports 88.1\% on MiniF2F-Test at pass@32 and Seed-Prover 99.6\%.

\textbf{Execution-Guided ATP.}
In execution-guided ATP, a prover interacts with the verification system iteratively during the proof search, generating proof steps conditioned on the internal proof state.
In neural theorem proving, this approach was pioneered by Holophrasm~\citep{whalen2016holophrasm} in Metamath~\citep{megill2019metamath}, using a UCT-based~\citep{kocsis2006uct} AND-OR tree search with a RNN sequence-to-sequence model for tactic enumeration and payoff prediction.
Later extensions include GPT-f~\citep{polu2020gptf} and Evariste~\citep{lample2022hypertree}, the latter also targeting Lean~3 in addition to Metamath.
ABEL~\citep{gloeckle2024abel} extended Evariste with incremental improvements and Lean~4 compatibility, but did not release the code or dataset.
AlphaProof~\citep{alphaproof} then achieved a significant improvement in performance, unfortunately also with closed code and data.

A separate line of open execution-guided provers performs linear best-first search over monolithic Lean proof states with a policy fine-tuned from a pretrained LLM: InternLM2.5-StepProver~\citep{wu2024internlm2}, BFS-Prover~\citep{xin2025bfsprover} with 72.95\% on MiniF2F-Test, BFS-Prover-V2~\citep{xin2025bfsproverv2} with 82.4\% and 86.1\% for its 7B and 32B models (95.1\% with an additional Gemini~2.5~Pro planner), and MPS-Prover~\citep{liang2025mpsprover} with 75.82\%.
In the terminology introduced above, these are execution-guided but linear provers: none of them factorizes proof states into independently searchable goals with node re-use and metavariable-coupling handling.
They release at most weights and inference code, and none releases training data or pipeline.
Being fine-tuned from Qwen2.5~\citep{qwen2025qwen25}, their reported budgets exclude the pre-training of the base model, on the order of $10^{12}$ TFLOPs, and their undisclosed pre-training corpora cannot be audited for benchmark contamination.
BFS-Prover also argues that, in its setting of a large pretrained policy and no learned critic, simple best-first search can outperform MCTS-style search; whether this carries over to factorized search with a trained critic is an open question that our release makes testable.

\textbf{Datasets.}
The standard library of Lean, Mathlib~\citep{mathlib2020}, offers more than 124k formal definitions and 255k formal proofs.
Together with numerous formalization projects created by the community, this offers a sizable human-written dataset for the supervised training of theorem provers.

However, execution-guided ATP approaches require additional data pre-processing to expose the intermediate proof states.
Such a dataset was first provided by LeanStep~\citep{han2021pact} in Lean 3 by extracting tactic invocation data from Mathlib.
LeanDojo~\citep{yang2023leandojo} extends this approach to Lean 4, additionally providing information about the premises used in each tactic application, necessary for training retrieval-augmented approaches.
Similar datasets are also offered by Pantograph~\citep{aniva2024pantograph} and lean-training-data~\citep{morrison2023lean-training-data}.
However, none of these datasets offers tactic proofs in a structured, simplified format, which is needed for NanoProof training.
Most importantly, they do not contain distance-to-proof-end information.

Additionally, Lean-Workbook~\citep{ying2024lean}, DeepSeek-Prover-V1~\citep{xin2024deepseek}, and NuminaMath-LEAN~\citep{wang2025kimina} provide large collections of automatically formalized proofs of various quality, the latter also offering a sizable human-formalized portion.

\textbf{Benchmarks.}
MiniF2F~\citep{zheng2021minif2f} offers 488 problems drawn from mathematical olympiad and undergraduate mathematics courses, formalized in parallel in Lean 3, Metamath, Isabelle, and HOL Light.
The objective is to generate valid formal proofs of the statements.
The benchmark was later ported to Lean 4 by \citet{yang2023leandojo}, with some formalization errors corrected by \citet{wu2024internlm2}.
We use MiniF2F as released by \citet{alphaproof}.

\section{NanoProof Automated Theorem Prover}\label{sec:nanoproof}

We describe the training and inference mechanism of NanoProof.
Section~\ref{sucsec:proof-search} mostly follows \citet{alphaproof}, while the following sections describe our implementation choices and additions; the LeanTree dataset and extraction tool that enable the supervised training of such a prover are described in Section~\ref{sec:leantree}.

\subsection{Proof Search Algorithm}\label{sucsec:proof-search}
\noindent\textbf{RL Environment.}
We formulate the proof search process as an RL environment.
At episode start, a~formal theorem statement is sampled from a~corpus (cf. Section~\ref{sec:experiments}).
At each step, the prover (\textit{agent}) observes the current state $s_t$ consisting of the list of open goals yet to be proven, and generates an action $a_t$, i.e., a Lean tactic.
The Lean verifier then implements the transition function from the pair $(s_t, a_t)$ to the list of next states $(s^1_{t+1}, \ldots, s^n_{t+1})$, potentially empty.
Note that this deviates from a~simple Markov Decision Process in that the agent proceeds independently into each next state.
Recall that for the goals in state $s_t$ to be proven, it is necessary to prove all goals in states $(s^1_{t+1}, \ldots, s^n_{t+1})$.

The episode ends when a valid action has been assigned to each state, yielding a proof tree.
The agent is then assigned a reward $r = -d$, where $d$ is the depth of the proof tree.
This incentivizes the agent to find short proofs that maximize branching, utilizing the factorization capability outlined above.

\noindent\textbf{MCTS Search.} Our proof search algorithm follows closely that of AlphaProof, and we leave detailed description to \citet{alphaproof}.
The prover consists of an \textit{actor} predicting which proof tactic to apply in the current proof state, and a \textit{critic} predicting future reward in the current proof branch, i.e., the negative depth of the proof sub-tree starting from the current state.
The algorithm maintains an AND-OR search tree that initially contains only the root node with the theorem to be proven and is then iteratively expanded by repeating three phases: selection, expansion and backpropagation, adapted from AlphaZero~\citep{silver2018alphazero} and Sampled MuZero~\citep{hubert2021sampledmu}.
At node expansion, generated tactics are immediately executed in Lean, and invalid ones are discarded.
Compared to sequential two-player games, the prover agent does not commit to intermediate actions and only performs a single search for each theorem.
For this reason, \citet{alphaproof} incorporates a progressive sampling mechanism, allowing new tactics to be sampled for internal nodes.

\noindent\textbf{Matchmaker.}
Theorems to be proven are sampled from a corpus with sampling weights determined by a simple heuristic \textit{matchmaker}.
After initial \texttt{trust\_count} attempts at each theorem, the matchmaker down-weights theorems that have been proven at least \texttt{trust\_count\_proved} in a row and theorems that have never been proved, thus prioritizing theorems at the edge of the prover's current ability.
Compared to \citet{alphaproof}, we additionally down-weight theorems for which a~single-tactic proof has been found, since no shorter proof can be discovered.

\subsection{Network Architecture}
The actor and critic are modeled using a single Transformer model~\citep{vaswani2017attention} based on nanochat~\citep{nanochat}.\footnote{\url{https://github.com/karpathy/nanochat}, MIT License}
Instead of utilizing dedicated actor and critic heads, we adopt the simpler approach of shared vocabulary by \citet{polu2020gptf}.
Specifically, the model input consists of a tokenized pretty-printed Lean state followed by a~special token indicating whether what follows is a~tactic or a~critic value.
During inference, tactics are decoded autoregressively.
In contrast, prediction of the critic value is done with only a single forward pass and weighted averaging of the value bins based on predicted token probabilities.

For tokenization, we use the GPT-2 tokenizer~\citep{gpt2}\footnote{\url{https://huggingface.co/openai-community/gpt2}, MIT License} extended with dedicated tokens for symbols that occur at least $1{,}000\times$ in Mathlib.
The range of allowed critic values is divided into discrete bins using dedicated tokens~\citep{farebrother2024stopregressing}.

\subsection{Training Pipeline}\label{subsec:training_pipeline}
As shown in Figure~\ref{fig:overview}, the prover network undergoes several supervised training phases before entering a self-improvement loop.
This preliminary phase is necessary because the generation of valid and diverse tactics is a challenging task in itself.
The supervised phases use increasingly targeted training data, starting with general math-related text and progressing through Lean source code to highly pre-processed human-authored Lean proofs.
This progression of data quality is mirrored in an increasingly limited quantity.

During the self-improvement loop, the prover continuously samples theorems from an autoformalized corpus and attempts to prove them, storing all successful proofs in a replay buffer.
Transitions sampled from the replay buffer are then used as further training data for the prover network.
In addition, a~given fraction of supervised fine-tuning data is mixed in with the collected data to increase training stability.
The system alternates between collecting $n_\text{samples}$ new proof transitions and performing $n_\text{updates}$ gradient updates.

As shown in Figure~\ref{fig:runtime_architecture}, the RL loop infrastructure features an inference server with simple first-available load balancing, a distributed trainer and detached Lean servers running on CPU nodes.

\subsection{Lean Integration}
LeanTree offers a~Python interface built on top of Lean REPL~\citep{lean4repl}.
After initiating a~proof search and applying a Lean tactic to a given proof state, LeanTree returns a~list of resulting sub-states.
The interface is suitable for tree search and backtracking as the interaction can be restarted from any intermediate proof state.
Additionally, given the high CPU requirements of Lean 4, LeanTree implements a dynamic pool of environments exposed over HTTP endpoint for deployment to CPU nodes during distributed training.

During our work, we encountered invalid proofs that Lean REPL accepted as valid; the details and a correction are described in Appendix~\ref{appendix:incremental_proof_verification}.
Note that all proofs found by NanoProof are independently verified by exporting them as standalone Lean code.

\subsection{Proof Simplification and Augmentation}
\label{subsec:proof_simplification_augmentation}

\noindent\textbf{Proof Simplification.}
We attempt to simplify the proofs produced by the proof search algorithm by pruning redundant tactics.
This is done both during experience collection to improve the quality of the data saved to the replay buffer and during final inference to enhance the readability of the generated proofs for humans.
In a representative RL run, our algorithm was able to simplify 59\% of proofs longer than one tactic.
An example of a simplified proof recorded during an experience collection phase is shown in Appendix~\ref{appendix:proof_simplification}.

\begin{figure*}[tb]
\centering

\begin{minipage}[t]{0.48\textwidth}
\vspace{0pt}
\linespread{1.22}\selectfont
\begin{algorithm}[H]
  \caption{Pruning of redundant tactics.}
  \label{alg:proof_simplification}
  \begin{algorithmic}
    \STATE {\bfseries Input:} proof tree $P$
    \medskip
    \STATE \textbf{procedure} $\textsc{PruneTactics}(P)$
    \medskip
    \FOR{any non-leaf $e = (s_t, s_{t+1})$ with action\rlap{~$a_t$}}
      \FOR{any $a_{t+1}$ applied in $s_{t+1}$}
        \STATE $P' \gets \text{contract}(P, e, a_t \leftarrow a_{t+1})$
        \IF{$\text{subtree}\_\text{valid}(s_t, P')$}
          \STATE \textbf{return} $\textsc{PruneTactics}(P')$
        \ENDIF
      \ENDFOR
    \ENDFOR
    \medskip
    \STATE \textbf{return} $P$
  \end{algorithmic}
\end{algorithm}
\end{minipage}
\hfill
\begin{minipage}[t]{0.48\textwidth}
\vspace{0pt}
\centering
\includegraphics[width=.85\linewidth]{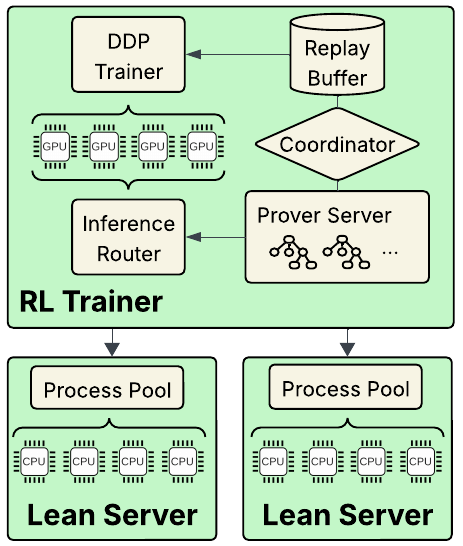}
\captionof{figure}{Runtime architecture of RL training.}
\label{fig:runtime_architecture}
\end{minipage}

\vspace{-1em}
\end{figure*}

The proof simplification algorithm is outlined in Listing~\ref{alg:proof_simplification}.
Given a generated proof, the simplification aims to extract a minimal sub-proof that still proves the original theorem.
Starting with a proof tree, we consider each non-leaf edge individually.
If contracting an edge leaves the proof valid, the edge is removed.
For a non-leaf edge $e = (s_t, s_{t+1})$ with tactic $a_t$, contraction is implemented by discarding $a_t$ (together with $s_{t+1}$), replacing it with any tactic $a_{t+1}$ applied in $s_{t+1}$.

When re-verifying the proof after contraction, it suffices to verify the modified proof sub-tree below $s_t$, since other branches and tactics on the path to root are not affected.
The sub-tree is verified by re-executing each tactic application in it, starting with the application of $a_{t+1}$.
When $a_t$ produces multiple factorized goals, i.e., it has more than one child node, we individually attempt to execute each of the sub-trees, and $e$ is deemed redundant when at least one execution succeeds.
After a node is removed, the whole simplification process is restarted because removing an edge might only be possible after a different edge is removed first.

\noindent\textbf{Proof Augmentation.}
When sampling a proof from the replay buffer, we apply a random augmentation to each of its proof edges, altering them while preserving their semantics.
First, all hypotheses and goals are randomly renamed with probability $p_{\text{rename}}$, applying the rename accordingly also in both the tactic and the other hypotheses.
Second, the hypotheses in each goal are randomly shuffled with probability $p_{\text{shuffle}}$.

\subsection{Cycle Detection}
We apply a cycle detection mechanism during node expansion.
When a tactic application results in a goal $G$ that is equivalent to a goal $G'$ already present on the path to root, the tactic is discarded.
Any final proof containing this tactic could equivalently be found by solving $G$ directly instead of descending to $G'$.
In a representative run, 18\% of otherwise valid tactics were rejected due to cycle detection.
Note that goal equivalence is checked at the syntactic level up to goal re-shuffling, goal renaming, and whitespace changes.

\subsection{Diverse Tactic Sampling}
We observe that the autoregressive tactic sampling often produces duplicate tactics, sometimes constraining node expansion to just a single action, particularly in the case of short one-token tactics.
In a representative run, 32\% of nodes had their action count reduced due to duplicated sampling.
We address this by forcing diverse sampling on the first tactic token.
Specifically, the first sampling step employs the Gumbel-Top-$k$ trick~\citep{gumbeltopk} to cap each possible token to \texttt{first\_token\_cap} occurrences.
Taking an average over last three saved checkpoints in the same run, this diverse tactic sampling increased performance on MiniF2F-Valid by 2.0 percentage points under 16 simulations, highlighting its importance for small compute budgets.

\section{Structured Proof Tree Dataset}\label{sec:leantree}

To enable supervised pretraining on existing human-written Lean proofs, factorized execution-guided provers require careful pre-processing of proofs into structured proof trees.
The challenge is twofold:
First, the Lean language understands a proof as a linear sequence of tactics, instead of a tree where different branches can be produced independently.
These tactics can be nested and combined with tactic combinators as well as other syntactic sugar.
Second, Lean proofs, as present in human-written repositories, are written in a compact, refined way to make them easy to read, maintain, and build on.
However, we seek a different proof style, one that unfolds the proof in a way amenable to search, i.e., in small incremental steps.

To address this challenge, we introduce a proof extraction tool and dataset, collectively called \texttt{LeanTree}, whose preliminary version was presented in \citet{kripner2025leantree}.
In each extracted proof tree, a node carries a single goal,\footnote{Note that, due to metavariable coupling, nodes occasionally contain more than one goal, which we detail in Section~\ref{sec:metavariable_dependent_goals}.}
and an edge corresponds to a tactic application operating on that goal.
The dataset contains $92{,}927$ extracted proof trees from a~recent version of Mathlib,\footnote{\texttt{leanprover/lean4:v4.27.0}, Apache-2.0 License} the standard library of human-written proofs in Lean.
Details about the structure of the dataset are provided in Appendix~\ref{appendix:schema}.
Importantly, each included tree is independently verified by executing it from the root down in topological order.
Because of technical challenges outlined in this section, not all tactic proofs can currently be verifiably converted.
Specifically, 9.53\% of tactic proofs in Mathlib were not included in LeanTree because the validity of all tactics in the extracted proof trees could not be verified.
We are continually working on perfecting this process.

The LeanTree dataset, together with its documentation and relevant metadata, is publicly available at \url{https://huggingface.co/datasets/ufal/leantree}.
The extraction tool and the Lean interface are available at \url{https://github.com/kripner/leantree}.

\begin{figure*}[t]
    \centering
    \includegraphics[width=\hsize]{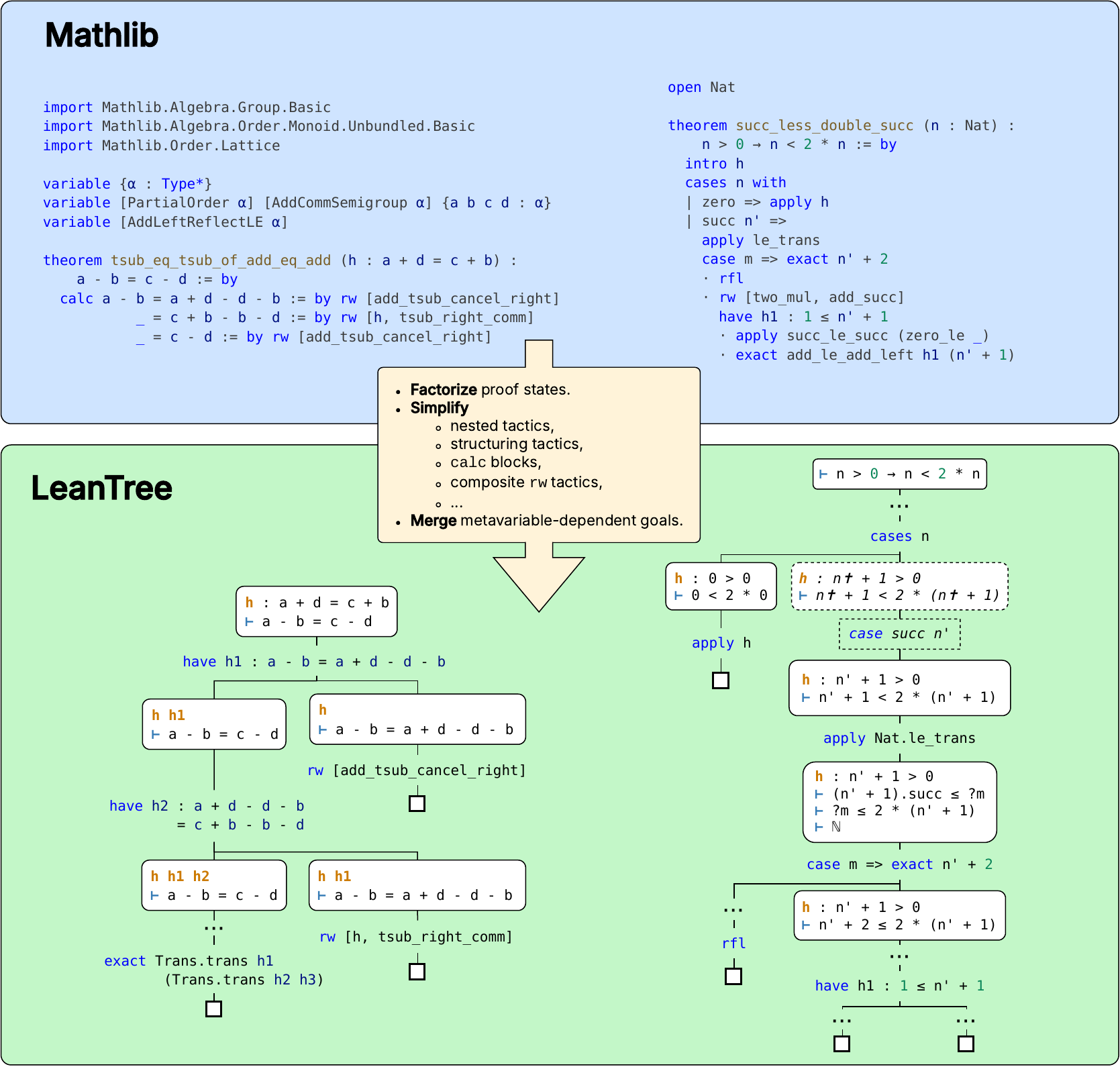}
    \caption{LeanTree extracts Lean proofs from a Lean file (top) and transforms them into structured proof trees (bottom). Synthetic tactics in the proof tree are marked using dashed border.}
    \label{fig:proof-tree-schematic}
\end{figure*}

\noindent\textbf{Proof Tree Building.}
To illustrate the challenges in building a proof tree, Figure~\ref{fig:proof-tree-schematic} shows two example Lean proofs and the resulting trees.
For the purposes of illustration, we do not utilize powerful tactics such as \texttt{linarith}, which would solve the trivial theorem nearly automatically.
Instead, we showcase branching, metavariable coupling, and the introduction of a new hypothesis.

Other Lean features not shown in Figure~\ref{fig:proof-tree-schematic} that further complicate the construction of proof trees include tactic combinators such as \texttt{try}, \texttt{iterate}, and \texttt{any\_goals}, where the set of affected goals is determined in runtime, constructs such as \texttt{all\_goals} and \texttt{<;>} that go against tree formulation due to operating on multiple goals, or tactics such as \texttt{switch} and \texttt{rotate\_left} that change the order of open goals.

We implement the proof extraction in three stages described in the following three subsections.
A~comparison of LeanTree with existing datasets and tools for training data extraction in Lean is given in Appendix~\ref{appendix:leantree_comparison}.

\subsection{Singleton Tree Construction}
The first stage of proof tree building transforms a Lean proof into a \textit{singleton tree}, where each node contains a single goal, by matching tactic applications with the goals they affect.
In general, a~tactic application transforms a list of open goals into an entirely new list of open goals without any additional guarantees, making this transformation nontrivial.
To build singleton trees, we integrate Lean REPL with an augmented version of the BetterParser module of PaperProof~\citep{kovsharov2024paperproof},\footnote{\url{https://github.com/Paper-Proof/paperproof}, MIT License} which solves an analogous problem, although originally developed for visualization purposes.

\subsection{Tactic Simplification}
\label{subsubsec:tactic_simplification}
Singleton trees are modified so that complex tactics are broken down into simpler ones, and each tactic is a valid self-contained proof step.
We now describe some of these modifications.

\textbf{Nested Tactics.}
Tactics parametrized by a term can themselves contain nested tactic blocks.
For example, consider the following tactic from the Complex Analysis module in Mathlib.

\vspace{-0.9\baselineskip}
\begin{equation}
    \begin{minipage}{0.9\linewidth}
        \begin{lstlisting}[language=lean]
exact ⟨by 
 rw [aeval_algHom_apply, hw, map_zero],
 rfl⟩
\end{lstlisting}
    \end{minipage}
    \label{lst:exact_with_by_block}
\end{equation}
\vspace{-0.7\baselineskip}

In Listing~\ref{lst:exact_with_by_block}, the \texttt{exact} tactic is parametrized by a term which contains a tactic block introduced using the \texttt{by} keyword.
We break down this complex proof step by masking the nested tactic block and instead spawning a child goal, yielding the following series of simpler proof steps.

\vspace{-0.9\baselineskip}
\begin{equation}
    \begin{minipage}{0.9\linewidth}
        \begin{lstlisting}[language=lean]
exact ⟨by sorry, rfl⟩
rw [aeval_algHom_apply]
rw [hw]
rw [map_zero]
\end{lstlisting}
    \end{minipage}
    \label{lst:exact_broken_down}
\end{equation}
\vspace{-0.7\baselineskip}

Note that the \texttt{rfl} tactic is not masked since it is used as a term instead of a tactic.

\textbf{\texttt{calc} Blocks.}
The \texttt{calc} tactic combines a sequence of relational proofs into a single transitive derivation.
We decompose such a block into one \texttt{have} tactic per step, followed by a final \texttt{exact} tactic that combines the intermediate lemmas.
An example is shown in Figure~\ref{fig:proof-tree-schematic}.

\textbf{Structuring Tactics.}
The \texttt{cases} and \texttt{induction} tactics introduce syntactic branching by partitioning a proof based on the value of an inductive type.
The tactic optionally contains a solving branch for each of the constructors of such a type.
We move the individual proof branches from the \texttt{cases} or \texttt{induction} tactic into their separate proof steps.
For example, consider the \texttt{cases} tactic in Figure~\ref{fig:proof-tree-schematic}, which spans 10 lines and should not be presented to a policy model as a training sample.

\textbf{Merged \texttt{rw} Tactics.}
In Lean, subsequent \texttt{rw} tactics can be merged into a single tactic, which is commonplace in human-written proofs.
We partition such composite tactics into individual \texttt{rw} tactics.

\subsection{Merging Metavariable-Dependent Goals}\label{sec:metavariable_dependent_goals}
To obtain the final proof tree, sibling nodes in a singleton tree are merged if they share a metavariable.
This corresponds to the fact that metavariable-dependent goals cannot be explored independently.

During this merging process, the entire proof tree is executed starting from the root state and following each tree edge, verifying that each proof branch ends in a proven state.
In this way, we ensure the correctness of the whole proof tree building process.
Proof trees that fail this verification are not included in the final dataset.

In Figure~\ref{fig:proof-tree-schematic}, observe that applying \texttt{Nat.le\_trans} yields three goals that are not split due to a shared metavariable \texttt{?m}.
However, once the metavariable is assigned using \texttt{exact}, the remaining two goals are now independent and can be split.

\begin{figure*}[t]
  \centering
  \begin{minipage}[t]{0.48\textwidth}
    \centering
    \includegraphics[width=1.05\linewidth]{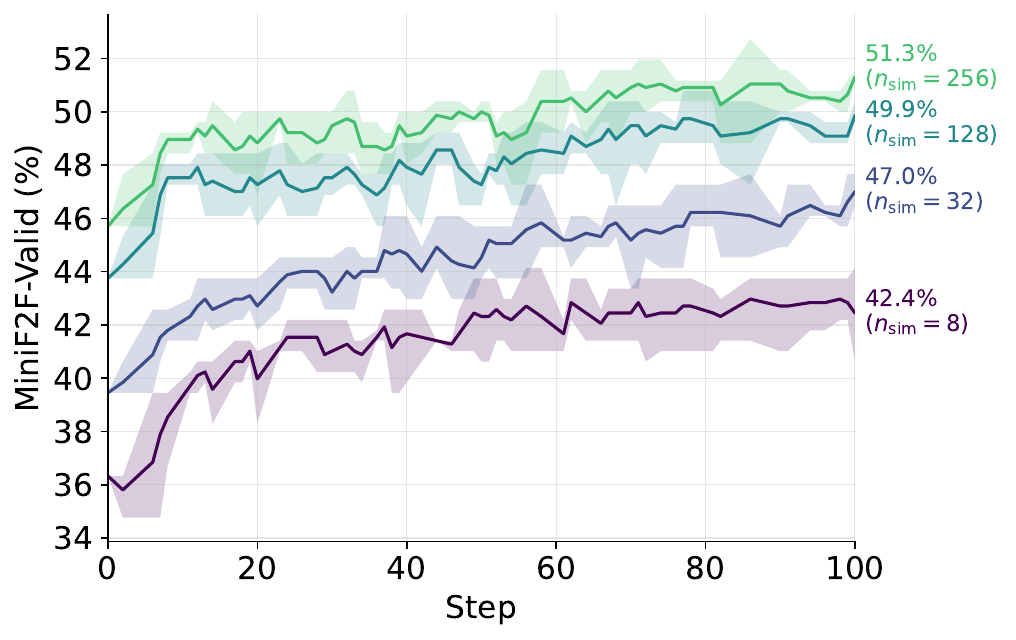}
    \caption{\textbf{Prover performance throughout RL training.} We show how NanoProof success rate on MiniF2F scales with RL training time and number of simulations $n_\text{sim}$. Averaged over three independent runs with min/max bands.}
    \label{fig:rl_progression}
  \end{minipage}\hfill
  \begin{minipage}[t]{0.48\textwidth}
    \centering
    \includegraphics[width=\linewidth]{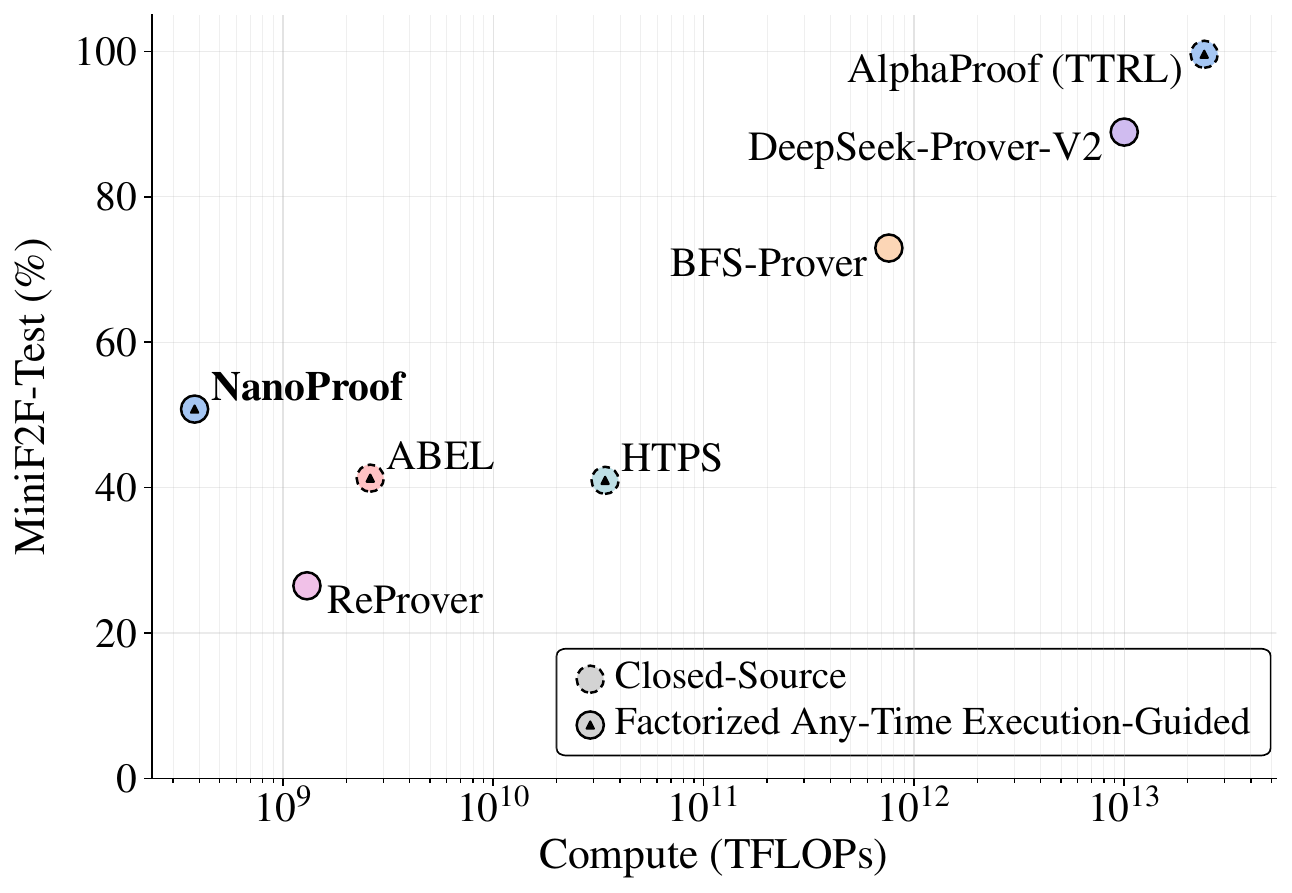}
    \caption{\textbf{FLOPs vs performance comparison.} We show ATP success rate on MiniF2F-Test relative to approximate total compute requirements. FLOPs are converted by us from reported hardware usage, or estimated where it is not disclosed (Appendix~\ref{appendix:compute_estimation}).}
    \label{fig:flops_vs_minif2f}
  \end{minipage}
  \vspace{-1em}
\end{figure*}

\section{Experiments}\label{sec:experiments}

We present an end-to-end training run of NanoProof.
The prover policy and critic network is first trained in a supervised fashion before entering into a self-improvement loop as described in Section~\ref{subsec:training_pipeline}.
We describe both of these phases below, with key hyperparameters listed in Appendix~\ref{appendix:hyperparameters}.
Hyperparameters were chosen manually after several experiments on MiniF2F-Valid, leaving MiniF2F-Test for final evaluation.

\textbf{Supervised Training.}
First, we \textit{base train} the network on the first 10B tokens of the Nemotron-CC-Math-4+ dataset~\citep{nemotron},\footnote{\url{https://huggingface.co/datasets/nvidia/Nemotron-CC-Math-v1}, NVIDIA Open Data License}
containing math and code related text.
Then, we \textit{fine-tune} the network on the 287k pre-processed proof steps from Mathlib~\citep{mathlib2020} contained in LeanTree, leaving the last 10\% of files (amounting to 29k proof steps) for validation and training on the remaining 258k.
Each proof step consists of a proof state, tactic, and distance to proof end, i.e., the corresponding sub-proof depth.
The data used during fine-tuning necessitate careful pre-processing described in Section~\ref{sec:leantree}.

We additionally experimented with inserting a \textit{mid-train} phase between base training and fine-tuning, consisting of additional training on 64M tokens of Lean code scraped from GitHub repositories listed in the LEAN-Github dataset~\citep{leangithub}.
However, we did not observe performance improvements and did not include mid-training in the final evaluations.

\textbf{Reinforcement Learning Loop.}
During RL training, we sample theorems from a corpus of $96{,}967$ autoformalized statements provided by the NuminaMath-LEAN dataset~\citep{wang2025kimina}.\footnote{\url{https://huggingface.co/datasets/AI-MO/NuminaMath-LEAN}, Apache-2.0 License}
Note that the dataset contains additional $5{,}529$ theorems which are incompatible with Lean~v4.27.

The RL training was run distributed over a node with eight GPUs and two CPU nodes, with the total hardware requirements detailed in Appendix~\ref{appendix:compute_estimation} (excluding preliminary experiments).

\subsection{Results}\label{subsec:results}

Figure~\ref{fig:rl_progression} shows the progression of NanoProof success rate on MiniF2F during RL training for different numbers of MCTS iterations, starting with the checkpoint after supervised training.
We perform three independent runs of RL training and report the mean score with min/max bands.
The runs are independent in terms of both the theorems sampled by the matchmaker and the tactics sampled during node expansion.

After 100 steps of RL training, or 45 hours, NanoProof achieves a mean performance of 51.3\% on MiniF2F-Valid over 3 runs (min: 51.2\%, max: 51.6\%) under the budget of 256 MCTS iterations.
A final checkpoint, selected from the three using validation split performance, achieves 50.8\% pass@16 on MiniF2F-Test under 512 iterations.

Our MiniF2F-Test score is contextualized in terms of the total compute budget in Figure~\ref{fig:flops_vs_minif2f} relative to selected existing provers.
Among the systems of the factorized any-time execution-guided class, NanoProof reaches a higher MiniF2F-Test score than HyperTree Proof Search and ABEL at roughly $90\times$ and $7\times$ less compute, and it is, to our knowledge, the only member of this class whose data, tooling, pipeline, and weights are all released.
Open-weight linear provers such as BFS-Prover report higher scores, but they start from a pretrained base model whose pre-training alone costs about three orders of magnitude more than our entire pipeline and cannot be reproduced; NanoProof instead reaches its score training from scratch, within a budget that any research group can rerun and modify.
The comparison is necessarily heterogeneous in Lean versions, pass@$k$ protocols, and search budgets, with compute figures converted by us from reported hardware usage, or estimated where it is not disclosed (Appendix~\ref{appendix:compute_estimation}).

\section{Conclusion}\label{sec:conclusion}
In this work, we identified and resolved a major roadblock in the open research of a class of automated theorem provers inspired by approaches such as AlphaProof and HyperTree Proof Search.
Furthermore, NanoProof, our end-to-end ATP system, reaches a higher MiniF2F-Test score than the previous systems of its class at a fraction of their compute, and is the only one that can be rebuilt from the released artifacts.
Along the way, we showcased a synthesis of available public datasets and filled in the missing pieces where necessary, particularly with the introduction of LeanTree, a dataset of structured proof trees extracted from Mathlib, and a tool for efficient distributed interaction with Lean.
We release all resulting artifacts as open-source.

\textbf{Limitations.}
We do not at this time possess enough computing resources to fully utilize the presented pipeline, leading to quantitative results that are sub-par compared to AlphaProof and the open-weight provers of Section~\ref{sec:related_work}, and to the absence of proper ablations: only diverse tactic sampling is ablated, while the other search-level techniques are reported through descriptive statistics only.
We do not claim methodological novelty in the search algorithm, which follows AlphaProof; our contribution lies in the data, tooling, and pipeline that make this class of prover reproducible.
The cross-system comparison is heterogeneous in Lean versions, pass@$k$ protocols, and search budgets, and the compute figures are converted from reported hardware usage or estimated where undisclosed.
NanoProof does not reach the level necessary for IMO or PutnamBench~\citep{tsoukalas2024putnambench} problems, and only MiniF2F is included in the evaluation.
Finally, 9.53\% of Mathlib proofs are not yet available in a~structured format in LeanTree.

\begin{acks}
This project was supported by the Charles University Grant Agency (GAUK), project no.~458326 and by the grant no.~25-18031S of the Czech Science Foundation (GAČR).
This research was partially supported by SVV project number 260~821.
We acknowledge VSB -- Technical University of Ostrava, IT4Innovations National Supercomputing Center, Czech Republic, for awarding this project access to the LUMI supercomputer, owned by the EuroHPC Joint Undertaking, hosted by CSC (Finland) and the LUMI consortium through the Ministry of Education, Youth and Sports of the Czech Republic through the e-INFRA CZ (grant ID: 90254).
This work was supported by the Ministry of Education, Youth and Sports of the Czech Republic through the e-INFRA CZ (ID:90254).
\end{acks}

\bibliographystyle{ACM-Reference-Format}
\bibliography{main}

\appendix
\clearpage

\section{Preliminaries: Lean and Proof Search}\label{sec:prelims}
In Lean, theorem statements correspond to types and proofs correspond to values.
Theorem proving then amounts to producing a term whose type is equal to the theorem type.
For example, consider the following theorem:

\vspace{-0.9\baselineskip}
\begin{equation}
\begin{minipage}{0.9\linewidth}
\begin{lstlisting}[language=lean]
theorem sub_zero (a b : ℕ) (h : b = 0)
  : a - b = a
\end{lstlisting}
\end{minipage}
\label{lst:sub_zero}
\end{equation}
\vspace{-0.7\baselineskip}

Proving Theorem~\ref{lst:sub_zero} consists in supplying a term of type \texttt{a - b = a} using the available free variables \texttt{a}, \texttt{b} of type $\mathbb{N}$ and \texttt{h} of type \texttt{b = 0}.
While such a term can be written down explicitly, it is common to utilize Lean's \textit{tactic mode} to structure the proof in a format more aligned with human-written proofs.

Upon entering tactic mode using the \texttt{by} keyword, Lean creates a \textit{metavariable} with the desired type \texttt{a - b = a} and no value assigned yet.
Metavariables can be thought of as holes in a proof that have to be filled in before the proof is considered finished.
This gives us an intermediate state in the proof construction, where the proof term is type-correct but still contains holes to be filled in later.
The still unassigned metavariables are called \textit{open goals}, so each goal has a target type and a set of available free variables called \textit{hypotheses}.
Because goals are metavariables, the two terms are often used interchangeably.

We then proceed by applying a \textit{tactic} to the list of open goals.
A tactic is a procedure that assigns values to one or more open goals, with each of the values possibly containing new unassigned metavariables.
In this way, a tactic invocation transforms the list of open goals into a new list of open goals.
Typically, a tactic only affects the \textit{main goal} (defined to be the first goal), either solving it or reducing it to one or more simpler sub-goals.
The proof is finished once there are no more open goals, i.e., every metavariable has been assigned.

\noindent\textbf{Proof State Factorization.}
We follow \citet{lample2022hypertree} in factorizing proof states into independent goals, which can be explored and proven individually.
Note that we temporarily disregard metavariable coupling, which is addressed later in this section.

\begin{figure*}[t]
\centering
\includegraphics[width=0.7\linewidth]{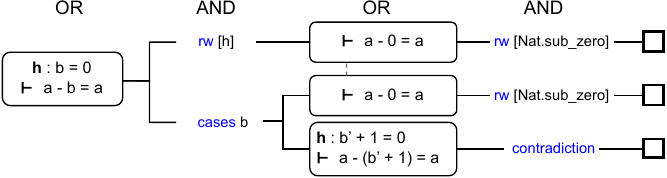}
\caption{
Example of an AND-OR proof tree for Theorem~\ref{lst:sub_zero}.
Nodes connected with a dashed line could potentially be merged and correspond to transpositions in games.
This is possible thanks to factorization.
}
\label{fig:and-or}
\end{figure*}

An example of potential state reuse is shown in Figure~\ref{fig:and-or}, where the same node is reached using two different search paths.

\vspace{-0.9\baselineskip}
\begin{equation}
\begin{minipage}{0.9\linewidth}
\begin{lstlisting}[language=lean]
theorem mul_eq_zero_iff (n m : ℕ)
  : n * m = 0 ↔ n = 0 ∨ m = 0
\end{lstlisting}
\end{minipage}
\label{lst:mul_eq_zero_iff}
\end{equation}
\vspace{-0.7\baselineskip}

A possible strategy to prove Theorem~\ref{lst:mul_eq_zero_iff} is to branch out based on the values of $n$ and $m$.
We achieve this by executing the tactic \texttt{cases n <;> cases m}.
Since each of the two numbers can be either zero or a successor of another number, the resulting proof state consists of four goals.
For example, one of the states is $(n' + 1) \cdot 0 = 0 \leftrightarrow n' + 1 = 0 \lor 0 = 0$, corresponding to the case where $n$ is a successor and $m$ is zero.

To complete the proof, all four goals must be closed.
In a non-factorized setting, the proof state consisting of four goals is assessed by the policy and critic models as a whole.
Since tactics typically affect only the main goal, the last goal can only be worked on once all others have been closed.
In contrast, factorized states allow the search algorithm to focus on the goals individually, in any order the algorithm deems fit.

\textbf{AND-OR Search.}
 The search of factorized states naturally yields an AND-OR tree, where a prover selects a tactic in an OR node and subsequently has to prove all the goals in an AND node. 
Figure~\ref{fig:and-or} shows an example in which choosing either the \texttt{rw} tactic or the \texttt{cases} tactic in the root leads to a proof.
For a more in-depth introduction to proof search in Lean, refer to \citep{aesop2023}.

\noindent\textbf{Metavariable Coupling.}
It is not always possible to fully factorize the proof state into individual goals that can be proven independently.
For this reason, nodes in the proof tree are allowed to contain more than one goal.
Specifically, this occurs when two goals share the same metavariable -- for example, applying the transitivity lemma \texttt{Nat.le\_trans} to the goal $2 \leq 5$ produces goals $2 \leq \texttt{?m}$ and $\texttt{?m} \leq 5$ sharing a metavariable \texttt{?m} whose value has not yet been decided.\footnote{A third goal of type $\mathbb{N}$ is created for technical reasons.}
Subsequently, choosing a value for \texttt{?m} in one of the goals affects the provability of the other goal, and therefore the two goals cannot be explored independently.
For a more in-depth explanation of this phenomenon, we recommend~\citet{aniva2024pantograph}.

\section{Hyperparameter Overview}\label{appendix:hyperparameters}
We give an overview of the most notable hyperparameters used during training and search in Table~\ref{tab:hyperparams}.
\begin{table*}[t]
\centering
\caption{Architectural, optimization and search hyperparameters.}
\label{tab:hyperparams}
\begin{tabular}{llc}
\toprule
Category & Hyperparameter & Value \\
\midrule
\multirow{8}{*}{Prover Network} & Total Parameters & 1B \\  %
& Number of Blocks & 26 \\
& $d_{model}$ & 1664 \\
& Attention Heads & 13 \\
& Feed Forward Widening Factor & 4 \\
& Vocabulary Size & $50{,}441$ \\
& RoPE Base ($\theta$) & $100{,}000$ \\
& Value Bins & $\{1, 2, \ldots, 64\}$ \\
\midrule
\multirow{9}{*}{Pre-Training} & Sequence Length & 768 \\
& Global Batch Size & $491{,}520$ \\
& AdamW Betas & $(0.8, 0.95)$ \\
& AdamW Epsilon & $1\times 10^{-10}$ \\
& Muon Momentum & 0.97 \\
& LR (embedding) & $0.3 \cdot 1/\sqrt{d_{model}/768}$ \\
& LR (unembedding) & $0.008 \cdot 1/\sqrt{d_{model}/768}$ \\
& LR (matrices) & $0.02$ \\
& LR Warmdown & last 65\% \\
\midrule
\multirow{1}{*}{Fine-Tuning} & Value Weight & 0.01 \\
\midrule
\multirow{6}{*}{RL Training} & Replay Buffer Max. Size & 250k transitions \\
& Supervised Data Fraction & 10\% \\
& Transitions per Step ($n_\text{samples}$) & 100 \\
& Examples per Update Step & 256 \\
& Updates per Step ($n_\text{updates}$) & 2 \\
& $n_\text{sim}$ during Experience Collection & 64 \\
\midrule
\multirow{10}{*}{Proof Search} & Maximum Tactic Length & 24 tokens \\
& Sampled Tactics & 6 \\
& Progressive Sampling $C$ & 0.1 \\
& Progressive Sampling $\alpha$ & 0.6 \\
& UCB value discount $\beta$ & 0.98 \\
& UCB prior temperature $\tau$ & 200 \\
& UCB exploration constant $c_{init}$ & 0.001 \\
& UCB exploration constant $c_{base}$ & 200 \\
& UCB AND-node prior multiplier $c_{AND}$ & 64 \\
& UCB unvisited children value penalty & 16 \\
\bottomrule
\end{tabular}
\end{table*}

\section{Proof Simplification}\label{appendix:proof_simplification}

We provide an example of a proof found during the RL training loop in Listing~\ref{lst:simplification_example_input} and its simplified version in Listing~\ref{lst:simplification_example_output}.
The latter was produced by the simplification algorithm described in Section~\ref{subsec:proof_simplification_augmentation}.
Only the simplified version is saved to the replay buffer during training.

\begin{equation}
    \begin{minipage}{0.9\linewidth}
        \begin{lstlisting}[language=lean]
theorem algebra_2complexrootspoly_xsqp49eqxp7itxpn7i (x : ℂ) :
    x ^ 2 + 49 = (x + 7 * Complex.I) * (x + -7 * Complex.I) := by
  subst_vars
  rw [← sub_eq_zero]
  rw [← sub_eq_zero]
  simp
  simp only [Real.cos_sq, Real.sin_sq, mul_comm, neg_mul]
  rw [sub_eq_add_neg]
  rw [mul_comm]
  rw [pow_two]
  rw [← sub_eq_zero]
  simp only [mul_comm]
  simp [sub_eq_add_neg]
  rw [add_comm]
  rw [← sub_eq_add_neg]
  rw [mul_comm]
  rw [mul_add]
  rw [sub_eq_add_neg]
  simp_all
  rw [add_comm]
  rw [← sub_eq_zero]
  rw [← neg_sub]
  rw [add_comm]
  rw [← neg_add]
  rw [sub_eq_add_neg]
  rw [← sub_self]
  rw [← neg_neg x]
  rw [sub_neg_eq_add]
  rw [neg_mul]
  rw [add_comm]
  ring_nf
  norm_num
\end{lstlisting}
    \end{minipage}
    \label{lst:simplification_example_input}
\end{equation}

\begin{equation}
    \begin{minipage}{0.9\linewidth}
        \begin{lstlisting}[language=lean]
theorem algebra_2complexrootspoly_xsqp49eqxp7itxpn7i' (x : ℂ) :
    x ^ 2 + 49 = (x + 7 * Complex.I) * (x + -7 * Complex.I) := by
  rw [← sub_eq_zero]
  ring_nf
  norm_num
\end{lstlisting}
    \end{minipage}
    \label{lst:simplification_example_output}
\end{equation}

\section{Structured Proof Tree Extraction}

\begin{figure*}[t]
    \centering
    \includegraphics[width=\hsize]{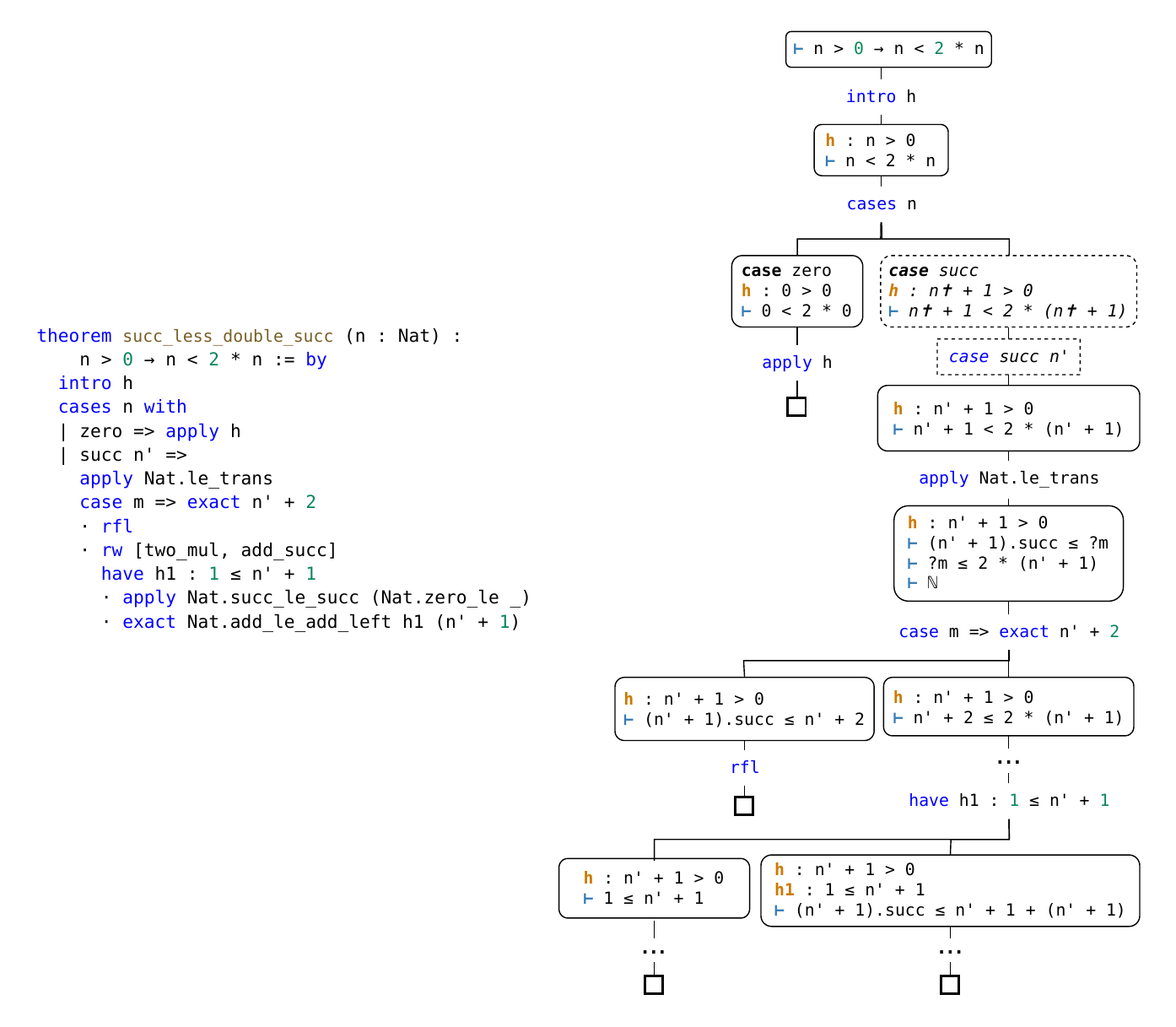}
    \caption{Example Lean theorem and its corresponding executable proof tree extracted using LeanTree. Some tactics and hypotheses are omitted.}
    \label{fig:proof-tree-full-A}
\end{figure*}

\begin{figure*}[t]
    \centering
    \includegraphics[width=\hsize]{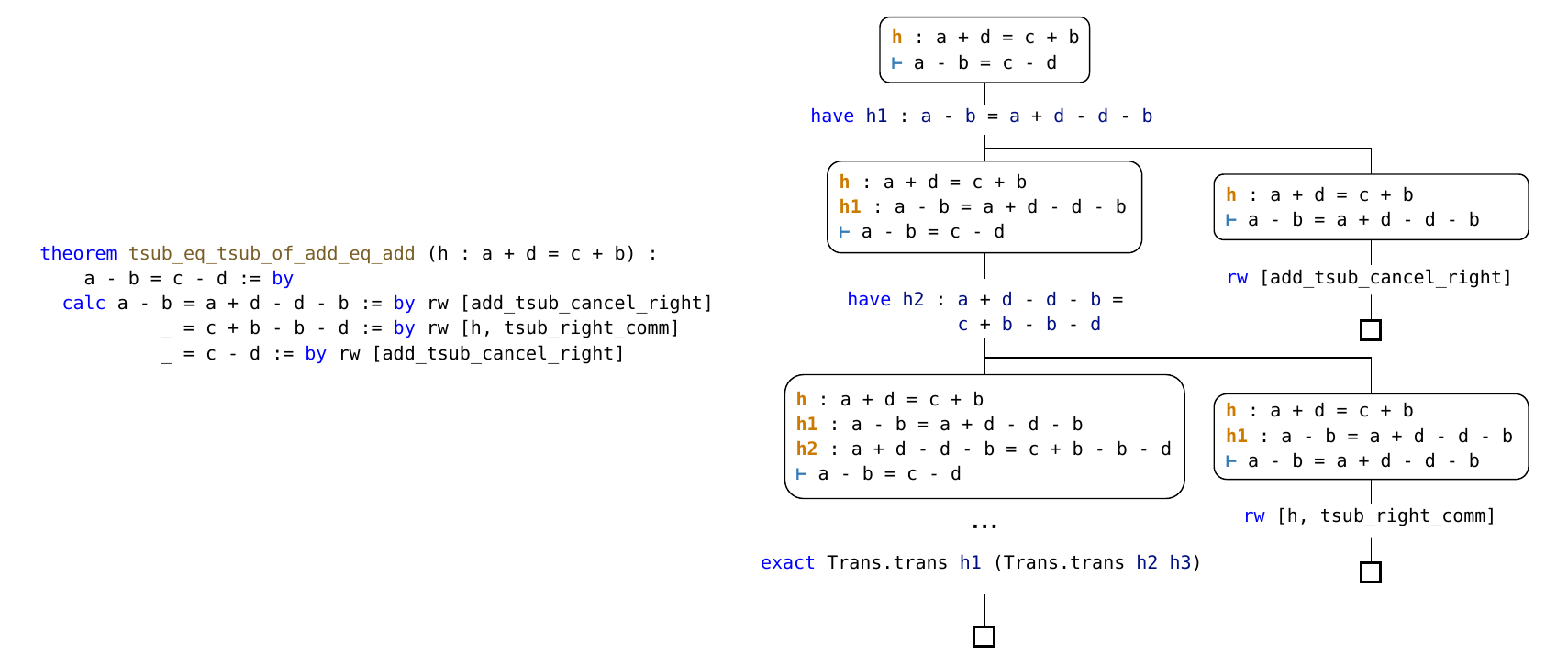}
    \caption{Example Lean theorem and its corresponding executable proof tree extracted using LeanTree. Some tactics and hypotheses are omitted.}
    \label{fig:proof-tree-full-B}
\end{figure*}

Figures \ref{fig:proof-tree-full-A} and \ref{fig:proof-tree-full-B} provide proof trees presented schematically in Figure~\ref{fig:proof-tree-schematic}.

\section{Incremental Proof Verification}
\label{appendix:incremental_proof_verification}

Originally, the Lean REPL did not utilize the Lean verification kernel to type-check assignments introduced by tactic applications, leading to incorrect proofs being accepted in some cases.
Our prover managed to exploit these inconsistencies, finding 12 incorrect proofs in the MiniF2F validation set that passed verification in the REPL when using the tactic mode.
One of these incorrect proofs is listed below.

\begin{minipage}{\linewidth}
\begin{lstlisting}[language=lean]
theorem mathd_algebra_422 (x : ℝ) (σ : Equiv ℝ ℝ) (h₀ : ∀ x, σ.1 x = 5 * x - 12)
    (h₁ : σ.1 (x + 1) = σ.2 x) : x = 47 / 24 := by
  rw [h₀] at h₁
  by_contra! h
  apply h.symm
  simpa using h _
\end{lstlisting}
\end{minipage}

Lean correctly rejects the above proof with the error: ``\textit{don't know how to synthesize placeholder}''.
However, when executing the proof in the tactic mode using the old version of Lean REPL, the proof passes as correct.

In a recent update,\footnote{Published on GitHub on Apr 10, 2025.} Lean REPL was extended to type-check the full proof term after each tactic application, mitigating all false positives in the verification process, at the cost of introducing some drawbacks detailed below.
We propose a different strategy: Instead of verifying the type-correctness of the whole proof term, type-check only the new assignments introduced by a tactic application.

To illustrate this difference, assume that we are searching for a proof of type $T$ and have already constructed a term $t : T$ which still contains unassigned metavariables of types $G_1, \ldots, G_n$.
Next, assume that we apply a tactic assigning terms $g_1 : G_1, \ldots, g_n : G_n$ to the metavariables in $t$.
The current strategy of Lean REPL is to type-check the whole assignment $t : T$, which contains the new terms $g_1, \ldots, g_n$ as subterms.

This leads to issues when the proof bifurcates into two branches.
For example, consider applying the tactic \texttt{have k := sorry} in the REPL's tactic mode.\footnote{This is the standard way to use the \texttt{have} tactic in the REPL -- see for example \url{https://github.com/leanprover-community/repl/blob/master/test/have_by_sorry.in}}
This corresponds to the assignment of the term \texttt{((fun k $\mapsto$ ?\_) sorry)} to the metavariable corresponding to the main goal, where \texttt{?\_} is a new unassigned metavariable.
Lean REPL then enables solving the \texttt{?\_} and \texttt{sorry} parts independently.
However, assignments made in one of these two branches are not propagated to the other one, meaning that, from the perspective of each branch, the root proof term will always contain at least one hole introduced by an unassigned metavariable or the \texttt{sorry} keyword.
This leads to the kernel always failing during a type check for the duration of this proof.
The end effect is that the current verification strategy leads to false negatives when the proof being verified contains at least one syntactic branching.

In contrast, the proposed strategy implemented in LeanTree type-checks only the new assignments $g_1 : G_1, \ldots, g_n : G_n$ introduced by a tactic application.
The type correctness of the whole proof term $t : T$ then follows transitively.
Most importantly, this new verification strategy mitigates the false negatives described in the previous paragraph.
Additionally, the new strategy also decreases the time complexity of verifying a proof from quadratic to linear since each sub-assignment is only verified once.

Note that all proofs found by our prover are independently verified directly using Lean, bypassing Lean REPL together with the new verification algorithm described in this section.

\section{Compute Estimation Methodology}\label{appendix:compute_estimation}

We estimate compute in FLOPs based on reported hardware usage, training time, and the following approximate conversion factors for BF16:

\begin{citemize}
  \item \textbf{NVIDIA H100 NVL}: 1671 TFLOPS
  \item \textbf{NVIDIA A100}: 312 TFLOPS
  \item \textbf{NVIDIA L40}: 242 TFLOPS
  \item \textbf{NVIDIA V100}: 125 TFLOPS
  \item \textbf{Google TPU v6e} (used in AlphaProof): 918 TFLOPS
  \item \textbf{AMD Instinct MI250}: 383 TFLOPS
\end{citemize}

Note that this form of compute budget estimation only provides rough estimates since the precise hardware characteristics and architectural details are not taken into consideration.

\subsection*{Per-Model Estimation Details}

\textbf{AlphaProof}~\citep{alphaproof}:
Computational budget as reported includes 10 TPUdays for base training, $100{,}000$ TPUdays for the autoformalization process, $80{,}000$ TPUdays for RL training, and 500 TPUdays per problem for TTRL.
These $302{,}010$ TPUdays (with test set size 244) correspond approximately to $2.4 \times 10^{13}$ TFLOPs.
Of this budget, $100{,}000$ TPUdays are spent on autoformalization.
NanoProof does not perform autoformalization but consumes the NuminaMath-LEAN corpus, whose generation cost is not disclosed and not included in our total.
Relative to NanoProof's $3.8 \times 10^8$ TFLOPs, the ratio is approximately $63{,}000\times$ for AlphaProof's total budget, $42{,}000\times$ when its autoformalization stage is excluded, and $17{,}000\times$ when counting only its RL stage; we therefore state the difference conservatively as more than four orders of magnitude.

\textbf{HyperTree Proof Search (HTPS)}~\citep{lample2022hypertree}: Compute budget as reported is 1260 A100-days for training with negligible inference cost.
This approximately corresponds to $3.4 \times 10^{10}$ TFLOPs.

\textbf{ABEL}~\citep{gloeckle2024abel}: Compute budget as reported is 96 A100-days for training with negligible inference cost.
This approximately corresponds to $2.6 \times 10^9$ TFLOPs.  

\textbf{DeepSeek-Prover-V2}~\citep{ren2025deepseek}:
Because the compute budget of DeepSeek-Prover-V2 has not been disclosed, we estimate it as approximately $10^{13}$ TFLOPs based on its architecture and training methodology:
it fine-tunes a 671B parameter model using RL over a large formal proof dataset with recursive self-play.
The estimate is conservatively aligned with models of similar size and RL complexity.

\textbf{ReProver}~\citep{yang2023leandojo}:
The training of retrieval and policy models requires 5 A100-days, corresponding to $1.3 \times 10^8$ TFLOPs.
Inference then requires 16 V100-days, corresponding to $1.7 \times 10^8$ TFLOPs.
However, the training was initialized from Google ByT5 Small~\citep{xue2022byt5}, which required an estimate of $10^9$ TFLOPs.
We thus report the total approximate budget as $1.3 \times 10^9$ TFLOPs.

\textbf{BFS-Prover}~\citep{xin2025bfsprover}:
The prover is fine-tuned from Qwen2.5-Math-7B, and its fine-tuning compute is not fully disclosed.
We therefore report a lower bound given by the pre-training cost of the base model alone, estimated as $6ND$ for $N$ parameters and $D$ training tokens, with 18T tokens for Qwen2.5~\citep{qwen2025qwen25}, giving approximately $7.6 \times 10^{11}$ TFLOPs.

\textbf{NanoProof:}
Here we describe one final run of NanoProof.
Supervised training, consisting of base training and fine-tuning, was performed on four A100-SXM4-40GB GPUs for a total of 16.3 hours.
RL training was performed on four AMD Instinct MI250 accelerators (each exposing two GPU devices, i.e., the eight GPUs of Section~\ref{sec:experiments}) for 45 hours together with two CPU nodes with comparatively negligible TFLOPS.
The final pass@16 performance was evaluated on four AMD Instinct MI250 GPUs for 11 hours.

This approximately corresponds to $7.3 \times 10^7$ TFLOPs for supervised training, $2.5 \times 10^8$ TFLOPs for RL, and $6.1 \times 10^7$ TFLOPs for final evaluation, for a total of $3.8 \times 10^8$ TFLOPs.
Note that this total includes the final evaluation ($6.1 \times 10^7$ TFLOPs, 16\%), while the figures for HyperTree Proof Search and ABEL cover training only.

\section{LeanTree Comparison to Existing Tools} \label{appendix:leantree_comparison}
Several tools have recently emerged for both programmatic interaction with Lean and machine learning data extraction from human-written Lean proofs.
In this section, we compare them to LeanTree.
We argue that LeanTree brings novel unique extraction capabilities.

\textbf{Evariste~\citep{lample2022hypertree}} corresponds most closely to LeanTree. Evariste enables factorizing proof state into individual goals, which can be explored independently.
In addition, Evariste detects metavariable coupling and does not split metavariable-dependent goals, similar to LeanTree.

However, the Evariste dataset has not been released and the tool itself is available only in a limited, nonexecutable format.\footnote{Specifically, the codebase states: ``code does NOT run out of the box as we removed references to Meta-internal systems''~\url{https://github.com/facebookresearch/Evariste}}
This makes it challenging to build on top of Evariste's contributions.

Importantly, Evariste was only implemented for Lean 3, which is now deprecated.
Since its core logic is implemented using C++, directly binding to Lean~3 internals, its migration to Lean 4 would be difficult.
In contrast, the core logic of LeanTree is implemented in Lean 4 itself, which is a common Lean 4 paradigm~\citep{moura2021lean}.
Furthermore, LeanTree builds on top of the semi-official Lean REPL project, and it is therefore set to benefit from future improvements by the community.

\textbf{LeanDojo~\citep{yang2023leandojo}} 
offers both an interaction module and a supervised data extraction module.
The extracted data also contains information about the premises used in each tactic application, enabling training of retrieval-augmented generation models.
The extracted datasets are available for both Lean 3 and Lean 4.

However, LeanDojo does not offer proof state factorization, tactic simplification, or proof tree building.
First, this means that a tactic can only be applied to a whole proof state consisting of a list of goals, yielding a new set of goals.
This prevents parallel search over independent goals.

Second, tactics in the LeanDojo dataset are left in the form in which they appear in the abstract syntax tree of a Lean source code.
This includes complex or nested tactics that span multiple lines.
For example, the \texttt{cases} tactic shown in Figure~\ref{fig:proof-tree-full-A} is left unaltered.
The user of the dataset can then choose to filter out such samples, losing valuable training data, or keep them in the training set, but increase the difficulty of the modeling task and reduce the granularity of the inference-time proof search.

Third, LeanDojo does not contain information about the sub-proof size and sub-proof depth for each node, and provides tactics in an unstructured format that is not conducive to proof tree building.

\textbf{Pantograph~\citep{aniva2024pantograph}}
In contrast to LeanDojo, Pantograph's interaction and data extraction modules offer proof state factorization, meaning that independent goals can be explored independently.
Pantograph also provides information about which goals are coupled via a shared metavariable.

However, tactic simplification and proof tree building are not supported by the data extraction module, leading to the same drawbacks as described for the case of LeanDojo.
Information about sub-proof size and sub-proof depth is also not provided.

In addition, the Pantograph API is not well suited for gradually building the current Lean context by interleaving definitions and theorems that can depend on any previously declared ones.
Instead, theorems are treated as independent units.

\section{Dataset Schema}\label{appendix:schema}

Each sample in the LeanTree dataset corresponds to a Lean file rather than just an individual theorem.
This is necessary to capture the structure of a real-world Lean project like Mathlib where a proof can depend on any definition located above it in the source file.

Each file in the dataset contains a list of theorems, and each theorem contains a list of all tactic proofs in its proof term.
Note that there can be more than one tactic proof for a theorem if its proof contains more than one non-nested \texttt{by}-blocks.
For each tactic proof, the dataset then contains a proof tree with nodes corresponding to factorized proof states and edges corresponding to tactic applications.
Every proof node contains also precomputed size and the depth of its sub-tree.

Additionally, the LeanTree dataset contains information about the surrounding context, namely the list of imported modules for each Lean file and the list of open namespaces for each theorem.
The correspondence between samples in the dataset and the underlying Lean repository is given by character offsets specifying the span of each theorem, proof, and tactic execution.

The JSON schema of the dataset is presented in Figure~\ref{fig:schema}.

\begin{figure*}[th]
\begin{lstlisting}[language=json]
<sample> ::= {
  "path": <string>,
  "imports": [<string>],
  "theorems": [<theorem>]
}

<theorem> ::= {
  "span": <span?>,
  "name": <string?>,
  "context": [<string>],
  "by_blocks": [<by_block>],
  "error": <string?>
}

<by_block> ::= {
  "span": <span>,
  "tree": {
    "nodes": [<proof_node>],
    "root_id": <string?>,
    "error": <string?>
  }
}

<proof_node> ::= {
  "id": <string>,
  "state": {
    "goals": [<goal>]
  },
  "tactic": <edge?>,
  "proof_size": <int?>,
  "proof_depth": <int?>
}

<edge> ::= {
  "tactic_string": <string>,
  "span": <span?>,
  "parent": <string>,
  "children": [<string>],
  "tactic_depends_on": [<string>?]
}

<goal> ::= {
  "type": <string>,
  "hypotheses": [<hypothesis>],
  "tag": <string>?
}

<hypothesis> ::= {
  "type": <string>,
  "user_name": <string>,
  "value": <string>?
}

<span> ::= {
  "start": <int>,
  "finish": <int>
}
\end{lstlisting}
\caption{JSON schema of a sample in the LeanTree dataset.}
\label{fig:schema}
\end{figure*}

\end{document}